\documentclass[5p,times]{elsarticle}
\journal{}%Future Generation Computer Systems (FGCS)}
\usepackage{amsmath,amssymb,amsfonts}
\usepackage{algorithmic}
\usepackage{graphicx}
\usepackage{textcomp}
\usepackage{times}
\usepackage{epsfig}
\usepackage{microtype}
\usepackage{booktabs} % for professional tables
\usepackage{lipsum}
\usepackage{physics}
\usepackage{wrapfig}
\usepackage{multirow}
\usepackage{enumitem}
\usepackage{tabularx}
\usepackage{colortbl} %use in the preamble
\usepackage{subcaption}
\usepackage{hyperref}

\begin{document}
    % Remove Elsevier stuff 
\makeatletter
\def\ps@pprintTitle{%
 \let\@oddhead\@empty
 \let\@evenhead\@empty
 \def\@oddfoot{}%
 \let\@evenfoot\@oddfoot}
\makeatother

% Main title of the paper
\title{Improving Novelty Detection using the Reconstructions of Nearest Neighbours}

% Address/affiliation
\address[1]{University of Amsterdam, Science Park 904, Amsterdam,1098XH,the Netherlands}
\address[2]{eScience Center, Science Park 140, Amsterdam, 1098XG, the Netherlands}
\address[3]{ASTRON, the Netherlands Institute for Radio Astronomy, Oude Hoogeveensedijk 4, Dwingeloo, 7991PD, the Netherlands}

\author[1]{Michael Mesarcik\corref{cor1}}
\ead{m.mesarcik@uva.nl}
\cortext[cor1]{Corresponding author}

\author[2]{Elena Ranguelova}
% Address/affiliation

% Corresponding author text
\author[3]{Albert-Jan Boonstra}

\author[1,2]{Rob V. van Nieuwpoort}

    \begin{abstract}
We show that using nearest neighbours in the latent space of autoencoders (AE) significantly improves performance of semi-supervised novelty detection in both single and multi-class contexts. Autoencoding methods detect novelty by learning to differentiate between the non-novel training class(es) and all other unseen classes. Our method harnesses a combination of the reconstructions of the nearest neighbours and the latent-neighbour distances of a given input's latent representation. We demonstrate that our nearest-latent-neighbours (NLN) algorithm is memory and time efficient, does not require significant data augmentation, nor is reliant on pretrained networks. Furthermore, we show that the NLN-algorithm is easily applicable to multiple datasets without modification. Additionally, the proposed algorithm is agnostic to autoencoder architecture and reconstruction error method.  We validate our method across several standard datasets for a variety of different autoencoding architectures such as vanilla, adversarial and variational autoencoders using either reconstruction, residual or feature consistent losses. The results show that the NLN algorithm grants up to a 17\% increase in Area Under the Receiver Operating Characteristics (AUROC) curve performance for the multi-class case and 8\% for single-class novelty detection.
\end{abstract}
    \maketitle
    \section{Introduction}
\label{sec:intro}
Novelty detection is an important field of research as identifying previously unknown behaviours in systems is critical for their maintenance and smooth operation. It is the procedure in which a model is able to identify new classes of data that it has not been exposed to before. Novelty detection is a far-reaching topic having been applied extensively in fields such as manufacturing \cite{Kim2012}, cyber-security \cite{Xu2018a}, biomedical analysis \cite{Schlegl2017, Baur}, astronomy \cite{Margalef-Bentabol2020} and many more \cite{Markou2003}. 

Novelty, anomaly, outlier, abnormality and out-of-distribution (OOD) detection are closely related topics~\cite{Ahmed2019}. The distinction between them is vague across variety of literature studies~\cite{Akcay,Perera2019,Pidhorskyi2018,Schlegl2017,Khan2010}. For clarity purposes, we consider novelty detection to be the overarching paradigm, since it makes contextual sense to have novel abnormalities/anomalies/outliers but the converse does not apply. 

Approaches for novelty and anomaly detection can be divided into a number of categories~\cite{Sipple2020, Chandola2009, Markou2003, Chalapathy2019}. In this work we exclusively focus on autoencoder-based novelty detection. As it offers a data agnostic method that does not rely on significant data augmentation~\cite{Li2021}, finding negative samples~\cite{Sohn2020,Fei2020} or pretraining on large labelled datasets~\cite{Bergman2020} such as ImageNet~\cite{Fei-Fei2010}. 

Autoencoders (AEs) are widely used as novelty detectors \cite{Akcay,Akcay2019,An2015,Perera2019,Pidhorskyi2018,Larsen2016, Li2020, Bergmann2019a, Liu2020a}. The underlying mechanism that governs the AE's detection abilities is that they are firstly trained on data without abnormal, anomalous or outlying samples. Then, during inference, the AE is exposed to novel samples which result in  higher errors thus enabling novelty detection. Methods such as mean-square-error (MSE) \cite{Larsen2016}, residual error \cite{Schlegl2017}, structural-similarity (SSIM) \cite{Bergmann2019} or feature consistency\cite{Lee2020} are used to calculate the pixel-wise difference. 

\begin{figure*}[htbp]
            \centering
           \includegraphics[width=\linewidth]{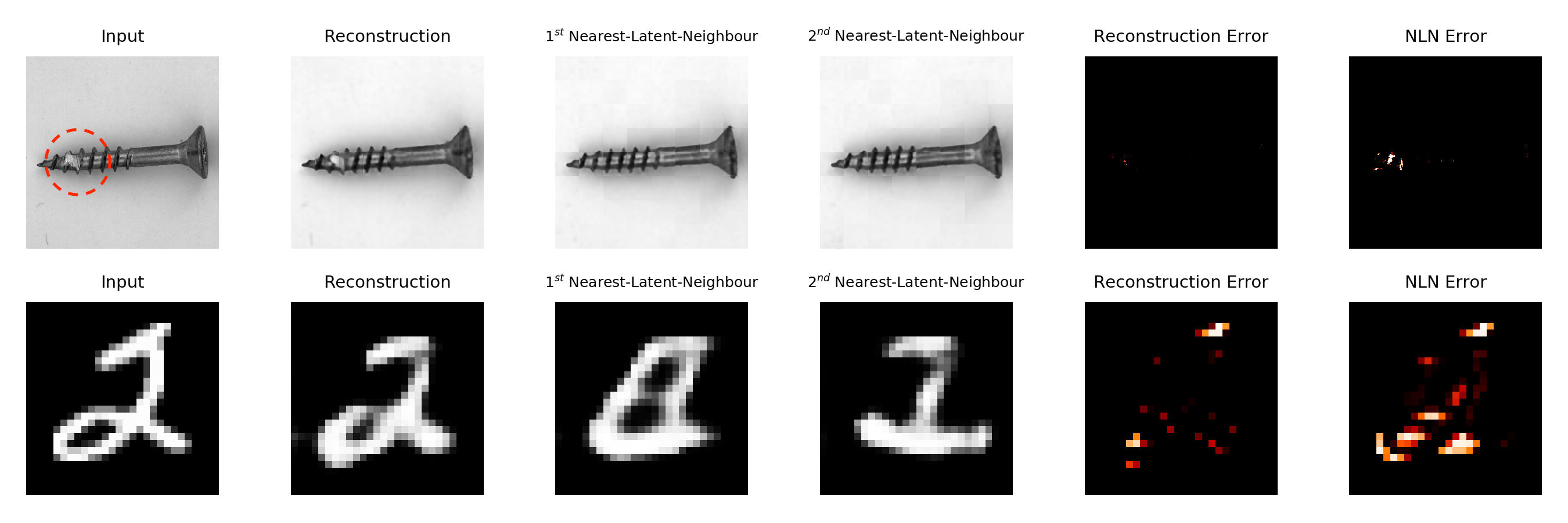}
            \caption{Comparison between the MSE and the Nearest-Latent-Neighbours (NLN) based error using a vanilla AE. The top row shows an AE trained on the non-anomalous \textit{screw} images of the MVTec-AD dataset, with the anomaly in the input circled in red, and the bottom row illustrates an AE trained on all MNIST digits except for "\textit{2}". The first column is the input to the AE, the second is the AE's output and the following two columns show the reconstructions of the input's NLNs. The final two columns show the difference between the MSE and the NLN-based error, where in this work maximising the error on novel classes effectively performs novelty detection. It is clear that the AE learns to reconstruct unseen classes whereas the reconstructions of the NLNs do not.} 
            \vspace{-1em}
            \label{fig:NLN_illustration}
\end{figure*}

A common problem with using autoencoding methods for novelty detection is that AEs can generalise to unseen classes thereby performing poorly as novelty detectors \cite{Hong2020}. In \cite{Perera2019}, this issue is addressed by placing a classifier in the training path of a multi-discriminator based autoencoder, which results in a fairly complicated and costly training procedure. On the contrary, we propose the Nearest-Latent-Neighbours (NLN) algorithm which uses the reconstructions of the nearest-neighbours in the latent space of autoencoders in-order to combat the aforementioned generalisation problem.  

Unlike existing nearest neighbours methods~\cite{Guo2018}, our NLN algorithm uses both the \emph{reconstruction error} between a given sample and its neighbours in the latent space as well as the average latent-distance to its neighbours. Figure~\ref{fig:NLN_illustration} illustrates how a vanilla autoencoder generalises to reconstruct unseen samples whereas the reconstructions of an input's nearest-latent-neighbours more closely resemble the non-novel training set thereby offering improved novelty detection. 

We evaluate the proposed method using the novelty detection framework described in \cite{Burlina2019} and prove it's effectiveness in a two-stage testing strategy. Firstly by comparing different architecture's performance with and without the use of our NLN algorithm. Secondly we compare our best performing model with the current state-of-the-art AEs. We show that NLN is competitive with the state-of-the-art methods across a number of datasets. 

In summary, the main contributions of this paper are: (1) a novel nearest-neighbour based algorithm that harnesses the reconstruction error of a given sample's nearest-latent-neighbours and their latent-neighbour distances. (2) The formulation of the NLN algorithm applied to a variety of autoencoding architectures using several different error calculation methods. (3) Improved performance to the state-of-the-art autoencoders using NLN, a fairly simple, cheap and intuitive method, across a number of standard datasets.

    \section{Background and Related Work}
    \label{sec:background}
    In novelty, anomaly, outlier, abnormality  and OOD detection one or more of the following steps are required for the detection of a novel, anomalous or outlying sample: (1) a model of the distribution of the (non-anomalous/non-novel) data. (2) A suitable measure of fitness describing whether a given sample lies within the modelled distribution. (3) A decision rule to determine whether the measure is above or below a threshold~\cite{Chandola2009}. 
    
    A critical distinction among all related work is whether supervised, semi-supervised,  or unsupervised methods have been used~\cite{Gu2019}. Supervised methods typically relate more strongly to anomaly detection scenarios, where both the normal and anomalous data classes are known \textit{a-priori}~\cite{Chandola2009}. However, supervision is not applicable in many settings, as anomalous classes are either underrepresented or just not known~\cite{Chandola2009}. In the unsupervised setup, we have no \textit{a-priori} information if the available data contains normal or abnormal samples~\cite{Gu2019}. 
    
    Semi-supervised methods are the most common in practice, as normal (non-anomalous, non-novel) data is most easily collected from most systems~\cite{Markou2003}. In this case, models are designed to represent the expected operating conditions of a system and any deviations from that are considered novel. These deviations may, in some cases, be considered anomalous, however this is dependent on the context of operation~\cite{Ahmed2019}. For the remainder of the paper we focus on semi-supervised methods, where we designate particular classes of a dataset as novel and all other as \textit{expected}. 
    
    \subsection{Reconstruction based novelty detection}
        Semi-supervised reconstruction based methods leverage the fact that a model trained on the \textit{normal} data cannot suitably reconstruct novel samples. In effect, the difference between the input and reconstructed output can be used as a novelty detector. Autoencoders (AE) are commonly used for reconstruction based novelty and anomaly detection~\cite{Akcay,Akcay2019,An2015,Perera2019,Pidhorskyi2018,Larsen2016, Li2020, Bergmann2019a, Liu2020a}. They operate by jointly learning latent representations and reconstructions of the training data. Once trained, a reconstruction error can be calculated between the input sample and the model's decoded output. These models achieve improved performance when regularising the latent space~\cite{An2015} using variational AEs (VAE)~\cite{Kingma2013a} or adversarial losses~\cite{Makhzani2015}.
        
        It has been demonstrated that reconstruction-error based methods alone are not particularly robust to noise, changing backgrounds and viewing angles~\cite{Ahmed2019}. In generative autoencoding models such as VAEs, reconstruction probability or attention-mechanisms are used to improve performance~\cite{An2015,Venkataramanan2020,Liu2020a}. Furthermore Generative Adversarial Networks (GANs)~\cite{Goodfellow2014} are used for reconstruction-error based anomaly detection~\cite{Schlegl2017, Zenati2018,Carrara2021}. Here the \textit{residual error} is calculated as the difference between the training and generated images using their intermediate representations provided by the discriminator.  More recently, self-supervised learning (SSL) has been applied to AEs and offers improved performance in novelty detection by using in-painting~\cite{Li2020} or position prediction~\cite{Yi2020} pretext tasks. 
        
        In our work we use the \textit{reconstructions} of a given sample's latent neighbours in conjunction with their latent-distances. We show that this offers performance increases across a variety of architectures and datasets. Additionally, we utilise several autoencoding models and show that the NLN algorithm offers performance improvements irrespective of architecture.

    \subsection{Statistical methods}
        Statistical methods typically focus on modelling a distribution of inliers through learning their distribution parameters~\cite{Pidhorskyi2018}. In effect, all expected/normal sample should lie in high density regions of the distribution, while outliers should have a low probability under the learnt distribution. Works such as One-class SVM~\cite{Erfani2016}, KNN~\cite{Gu2019} and isolation forests~\cite{TonyLiu2008} have been shown to be suitable anomaly detectors applied in this context.  
        
        Furthermore, \cite{Burlina2019} demonstrates that using discriminative measures in the latent space of AEs improves accuracy over reconstruction error. Here, discriminative novelty measures such as One-class SVM and   Local Outlier Factor (LOF)~\cite{Mudinas2020} are applied to the latent space of AEs.

        In our work, we propose a hybrid approach, where we combine a nearest neighbours based approach, that is typically used in distance-based anomaly detection, with a reconstruction error-based approach. This is done by considering the reconstruction error between neighbouring points in the latent space with some input query. We show that our work enables robustness and improvements to existing state-of-the-art research. 
            
    \subsection{Single and Mutli-class novelty detection}
        In the context of deep learning, one-class (single-class) novelty detection~\cite{Perera2019,Abati2019,Sabokrou2018,Pidhorskyi2018,Merelli2004,Zenati2018a} is the paradigm where a single class is considered normal and all other classes are novel. In practice, a model is trained on a dataset consisting of only a single class and during inference the novelty detector is exposed to all classes and should identify all unseen classes as novel. 
        
        For multi-class novelty detection, multiple classes are considered inliers and a single class is considered novel~\cite{Akcay,Akcay2019,Lee2020,Zenati2018}. This is an inherently more challenging evaluation framework as the model should be able to generalise to multiple classes and still be capable of detecting novel samples. In this work we evaluate our NLN-enabled models in both a Multiple-Inlier-Single-Outlier (MISO)  and Single-Inlier-Multiple-Outlier (SIMO) contexts as defined by~\cite{Burlina2019}.

    \section{NLN: Nearest Latent Neighbours}
\label{sec:proposed_method}
    Here we present our novelty detection framework for autoencoders. We show that using a simple addition to existing autoencoding architectures we can significantly increase their novelty detection performance.
    
    \subsection{Motivation}
        In \cite{Perera2019} and \cite{Hong2020} the limitations of using AEs for one-class novelty detection are demonstrated. They show that when an AE is trained on the relatively complex \textit{8}-class from the MNIST dataset~\cite{LeCun2010}, the AE is able to implicitly learn the representations of digit classes such as the \textit{1}, \textit{3}, \textit{6} and \textit{7}. In effect, reconstruction-based novelty detectors are prone to misidentify these implicitly learnt classes.  
        
        In order to solve this problem, \cite{Perera2019} propose placing a classifier in the training path of a multi-discriminator-based AE to decrease the training signal for the reconstructions of implicitly learnt novel classes. Conversely, we show that if we consider both the distance to, and the reconstruction of, a given sample's nearest latent neighbours we can effectively mitigate this issue, as demonstrated in Figure~\ref{fig:NLN_illustration}.  
        
        Furthermore, we motivate our focus on AEs for novelty detection as they are applicable to a variety of datasets without significant augmentation \cite{Li2021,Fei2020}, do not need pretraining on large labelled datasets \cite{Bergman2020, Reiss2020} and require far fewer network parameters \cite{Bergmann2021}. Additionally, their structure provides segmentation maps for free without the need of many small patches \cite{Yi2020} that result in a significantly more expensive KNN search \footnote{In \cite{Yi2020} 2 KNN searches are performed on patches sizes of 32 and 64, whereas we only need a single lookup for patches of size 128}  or additional networks for segmentation~\cite{Li2021}.
        
    \subsection{Problem formulation and approach}
        Considering an autoencoding model with encoder, $f$, and  decoder, $g$, then 
        \begin{equation}
            f(\vb{x};\theta_f) = \vb{z}, \quad f(\vb{x};\theta_f): \mathbb{R}^{p} \rightarrow \mathbb{R}^{l}
        \end{equation}
        where $\vb{x}$ is the input, $\vb{z}$ is the input's latent representation and $\theta_f$ are the parameters of the encoder. Additionally, $\mathbb{R}^p$ is the $p$-dimensional image-space and $\mathbb{R}^l$ is the $l$-dimensional latent-space. Now consider the decoder with an input, $\vb{z}$, and a reconstructed output $\hat{\vb{x}}$ such that
        \begin{equation}
            g(\vb{z};\theta_g) = \hat{\vb{x}}, \quad g(\vb{z};\theta_g): \mathbb{R}^{l} \rightarrow \mathbb{R}^{p}
        \end{equation}
        where $\theta_g$ are the decoder's parameters, such that the decoder maps from the $l$-dimensional latent space to the $p$-dimensional image space. The encoder and decoder pair is trained in an end-to-end manner using a loss function such as Mean-Square-Error (MSE) or Binary-Cross-Entropy(BCE). Once trained, the AE's novelty score ($\eta$) is computed for the $i^{\text{th}}$ sample using 
        \begin{equation}
            \label{eq:orig}
            \eta = \dfrac{1}{NM}\sum^N \sum^M|\vb{x}_i[n,m] - \hat{\vb{x}}_i[n,m] | \; 
        \end{equation}
        
        where $n$ and $m$ are the pixel-indexes for an image of size $N\times M$. This score is typically thresholded in order to determine whether a sample is novel and the threshold is calculated using AUROC-based methods that are explained in more detail in Section~\ref{sec:experiments}.

\begin{figure}[htbp]
                \centering
                \includegraphics[width=\linewidth]{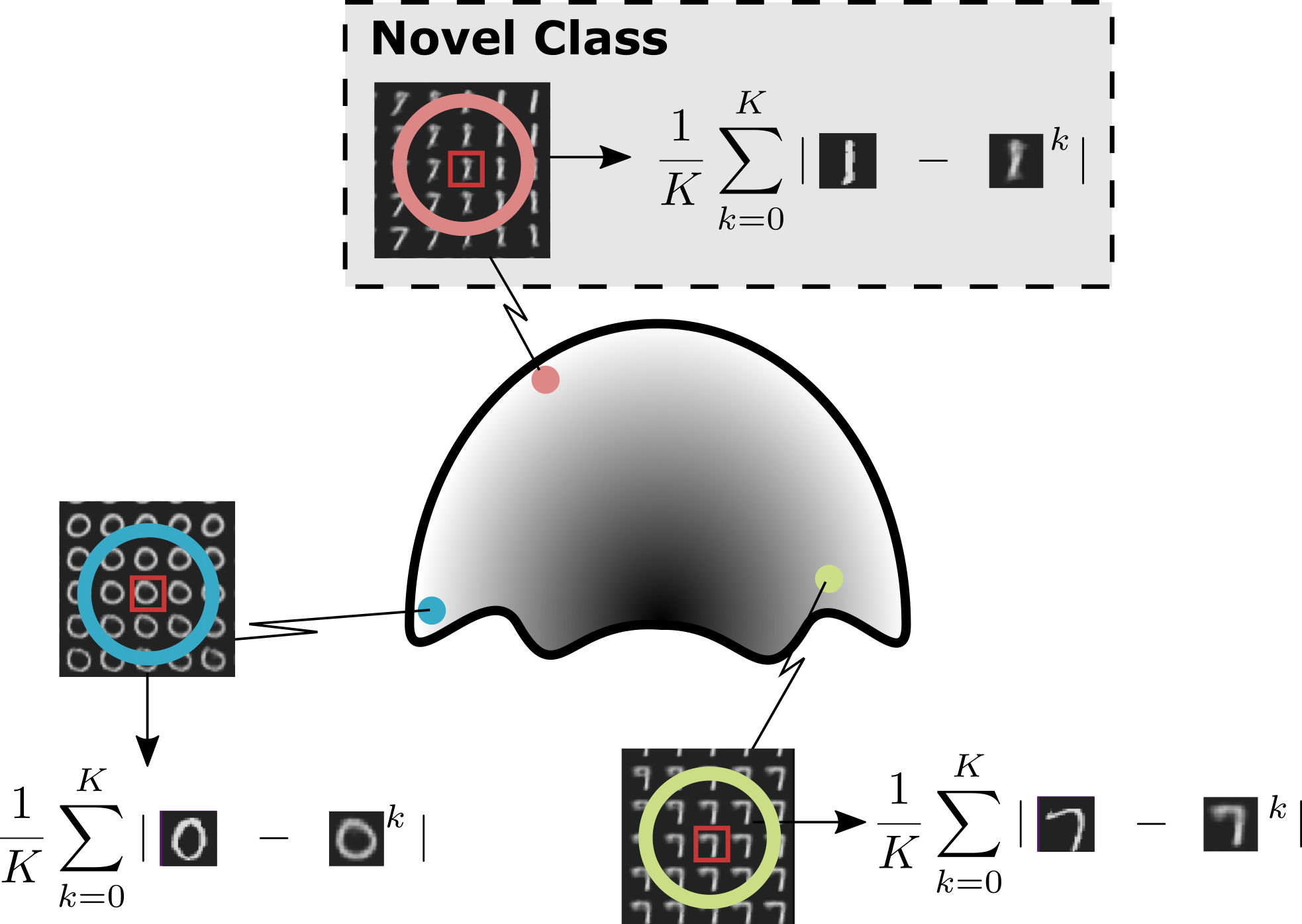}
                \caption{Illustration of the learnt MNIST data manifold trained without the class of \textit{1}'s (the class of \textit{1}'s are novel). The closely-connected regions of the novel class of \textit{1}'s contain dissimilar digits resembling \textit{7} and \textit{9} whereas the non-novel classes consisting of \textit{0} and \textit{7} do not.}
                \label{fig:manifold}
            \end{figure}

        In order to motivate our use of nearest-neighbours, we assume that the high-dimensional training data is concentrated on a low-dimensional data manifold in $\mathcal{R}^l$ that we attempt to learn using autoencoder~\cite{Goodfellow2015}. The learnt manifold is illustrated in Figure~\ref{fig:manifold}. Here we demonstrate that closely-connected regions on the learnt manifold contain points similar to non-anomalous inputs and dissimilar to those which are novel. We exploit this fact to improve the anomaly score robustness by including the nearest-latent neighbours into the reconstruction error. This is done by including the neighbours of the $i^{\text{th}}$ test sample in the latent space $\mathbb{R}^l$ in the calculation of the novelty score ($\eta_{\text{nln}}$). Such that   
        
        \begin{equation}
            \label{eq:nln}
            \begin{split}
             \eta_{\text{nln}} = &\dfrac{\alpha}{KNM} \sum^K\sum^N \sum^M |\vb{x}_{i}[n,m] - g(\vb{z}_i^k;\theta_g)[n,m] | \\   
             + &\dfrac{1-\alpha}{K} \sum^K |\vb{z}_i  - \vb{z}_i^k | \; 
            \end{split}
        \end{equation}
        
        where $k$ is the neighbour index such that $\vb{z}^k_i$ is $\vb{z}_i$'s nearest neighbours in the latent space. $K$ is the maximum number of latent neighbours and $\alpha$ is the hyper-parameter ($\in [0,1]$) used to tune the contribution of latent-space and image-space based distances respectively. It must be noted that Equation~\ref{eq:nln} shows the critical difference between the \cite{Bergman2020,Guo2018} and our work. We propose using the reconstruction error in the image space, $\mathbb{R}^p$, whereas they only use the difference of extracted feature vectors in $\mathbb{R}^l$.

       \subsubsection{Discriminative Considerations}
           Discriminative autoencoding models use discriminators in the training of autoencoders. This is done to either  improve the \textit{realism} of the AE's outputs or to regularise the latent space to a prior distribution. In this work we focus on the former case. Given a discriminator $d_{\vb{x}}$, trained on inputs $\vb{x}$ and $\hat{\vb{x}} = g(\vb{z};\theta_g)$ then 
            \begin{equation}
              d_{\vb{x}}(\vb{x};\theta_{d_{\vb{x}}}): \mathbb{R}^{m} \rightarrow [0,1].
            \end{equation}
            Where the discriminator on $\vb{x}$ maps between the image space and a value on the interval between 0 and 1. It returns \textit{0} or \textit{1} based on whether the sample $\vb{x}$ is taken from the training set or if it is generated by the decoder, $g$. The discriminator's training objective is stated as \cite{Goodfellow2014} 
            
            \begin{equation}
                    \mathcal{L}_{\text{disc}}  =   \mathbb{E} [ \log ( d_{\vb{x}}(\vb{x}) )]  + \mathbb{E} [ \log(1-d_{\vb{x}}( \hat{\vb{x}}  )) ]
            \end{equation}
           In addition to improving the regularisation, discriminators can also be used for novelty detection. Novelty is calculated through the difference between the representations of a sample $\vb{x}_i$, and its respective decoded output $\hat{\vb{x}}_i$, from an intermediate layer, $q$, of $d_{\vb{x}}$. This is also referred to as the residual error \cite{Schlegl2017} and we include the nearest-latent-neighbours by 
            \begin{equation}
                \label{eq:nln_residual}
                \begin{split}
                \eta_{\text{res}} = &\dfrac{\alpha}{HK} \sum^H\sum^K |q(\vb{x}_{i})[h] - q(g(\vb{z}_i^k;\theta_g))[h] | \\
                + &\dfrac{1-\alpha}{K} \sum^K |\vb{z}_i  - \vb{z}_i^k |
                \end{split}
            \end{equation}
            where $h$ in an index of the output from an intermediate layer $q$ with size $H$. 
            
            \subsubsection{Feature Consistency}
            It has been shown in~\cite{Akcay} that adding an additional encoder in the training path of the autoencoder improves performance. This paradigm is referred to as feature consistency \cite{Lee2020} and can be integrated in our nearest-latent-neighbours method by
            \begin{equation}
                \label{eq:nln_con}
                \begin{split}
                 \eta_{\text{con}} = &\dfrac{\alpha}{LK} \sum^L \sum^K | f(\vb{x}_i;\theta_{f})[l] - f_{\text{con}}(\hat{\vb{x}}^k_i;\theta_{f_{\text{con}}})[l] | \\   
                 + &\dfrac{1-\alpha}{K} \sum^K |\vb{z}_i  - \vb{z}_i^k |.
                \end{split}
            \end{equation}
            Where $f_{\text{con}}$ is the additional encoder that takes $\hat{\vb{x}}$ as an input, with parameters, $\theta_{f_{\text{con}}}$. Furthermore, $L$ is the latent space dimensionality, which is maintained between the first encoder, $f$, and the second encoder, $f_{\text{con}}$ and is indexed by $l$. The encoder is trained jointly with the rest of the discriminative autoencoder as described in \cite{Akcay}.

        \begin{figure*}[htbp]
             \centering
             \includegraphics[width=\textwidth]{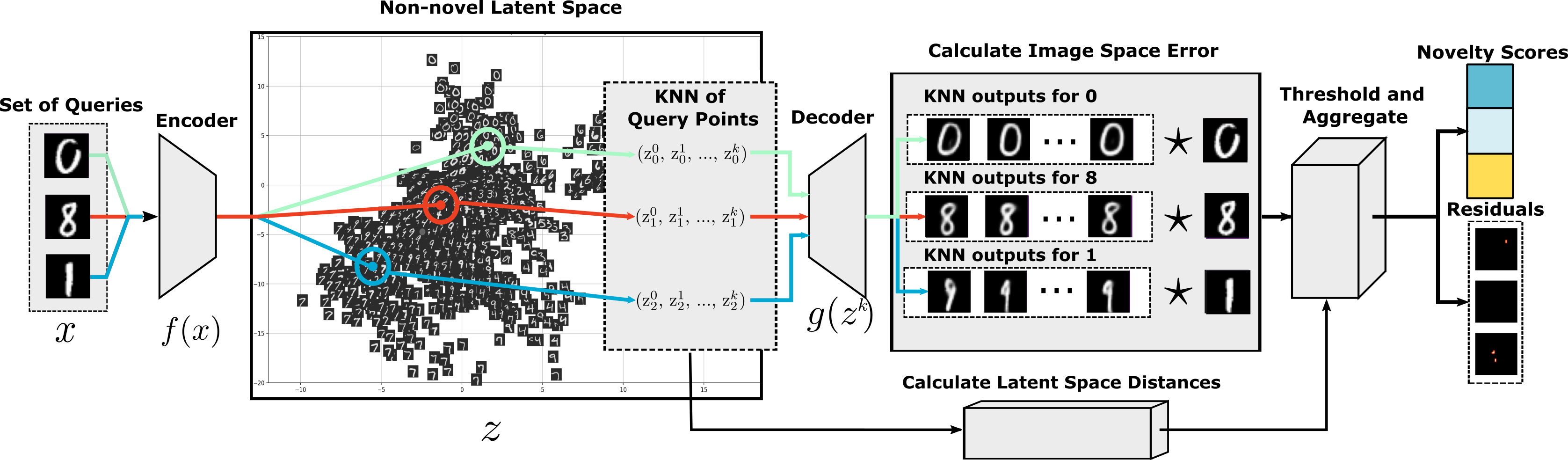}
             \caption{An illustration of our NLN algorithm, with three different samples from MNIST used as inputs to an autoencoding model, trained without the class of \textit{1}'s (i.e., \textit{1}'s are novel). Each of the three input's latent representations are found and their nearest neighbours (relative to the training set) are computed. The $\star$ represents the chosen error calculation operator. Finally, the error is thresholded and the residual-error maps and novelty scores are produced. }
             \label{fig:model}
        \end{figure*} 
            
    \subsection{The NLN algorithm} 
        Our work concerns the integration of the NLN technique into existing autoencoding models. For this reason we explain three different modes of operation for three different novelty scores. In the first case, a vanilla autoencoding model is used with a standard reconstruction error, as shown in Equation~\ref{eq:nln}. The second uses the autoencoding architecture in \cite{Akcay} and the feature consistency error in Equation~\ref{eq:nln_con}. Finally, the third makes use of a discriminative autoencoding architecture and use of the residual error in Equation~\ref{eq:nln_residual}.
        
        In all cases, an autoencoding model is first trained on a dataset with some novel class(es) removed. During testing, a sample is randomly chosen (which may be novel or not) and is input into the encoder.
        Then the nearest neighbours of the encoded sample are found in the latent space generated by the training data. This process is represented by the left-most half of Figure~\ref{fig:model}.
        
        In the first mode of operation, the error is computed between the test sample and both the decodings and positions of its latent-neighbours in the non-novel latent space. Whereas when discriminative methods are used, the error is computed between the intermediate representation from the discriminator $d_{\vb{x}}$ of the test sample and all its decoded latent neighbours in the training data. In the feature-consistent case, the error is computed between the encoding via $f_{\text{con}}$ of the given sample and all its nearest-latent-neighbours in the training data. In Figure~\ref{fig:model} these three operations are represented by the $\star$ operator.
        
        When performing novelty detection, one of the three methods' errors are aggregated over all neighbours and normalised after which they are added to the aggregated and normalised latent-neighbour distance vector. Then they are thresholded to result in an anomaly score and a segmentation map. The threshold is determined by the AUROC method described in Section~\ref{sec:experiments}. This methodology is illustrated in the right half of Figure~\ref{fig:model}.

    \section{Experiments}
    \label{sec:experiments}
    We evaluate our method\footnote{Source code available at: \url{https://github.com/mesarcik/NLN}} experimentally in both multi-class and single-class novelty detection contexts as outlined in \cite{Burlina2019}. Furthermore, we compare our best performing NLN-enabled autoencoder using both pixel-level and image-level anomaly detection metrics on the MVTec-AD dataset with state-of-the-art autoencoders. 
    
      \begin{figure*}[h!]
            \centering
            \begin{subfigure}[b]{0.49\textwidth}
                \centering
                \includegraphics[width=\linewidth,trim={3cm 0 2cm 0},clip]{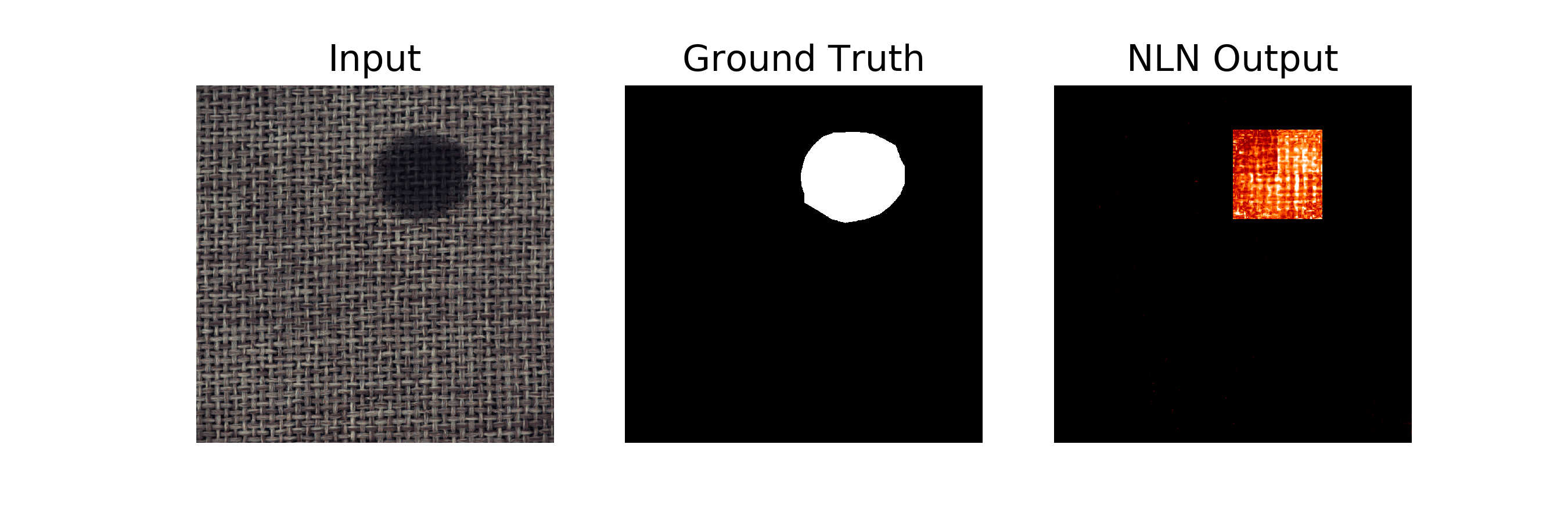}
                \caption{Carpet}
                \label{fig:carpet}
            \end{subfigure}
            \hfill
            \begin{subfigure}[b]{0.49\textwidth}
                \centering
                \includegraphics[width=\linewidth,trim={3cm 0 2cm 0},clip]{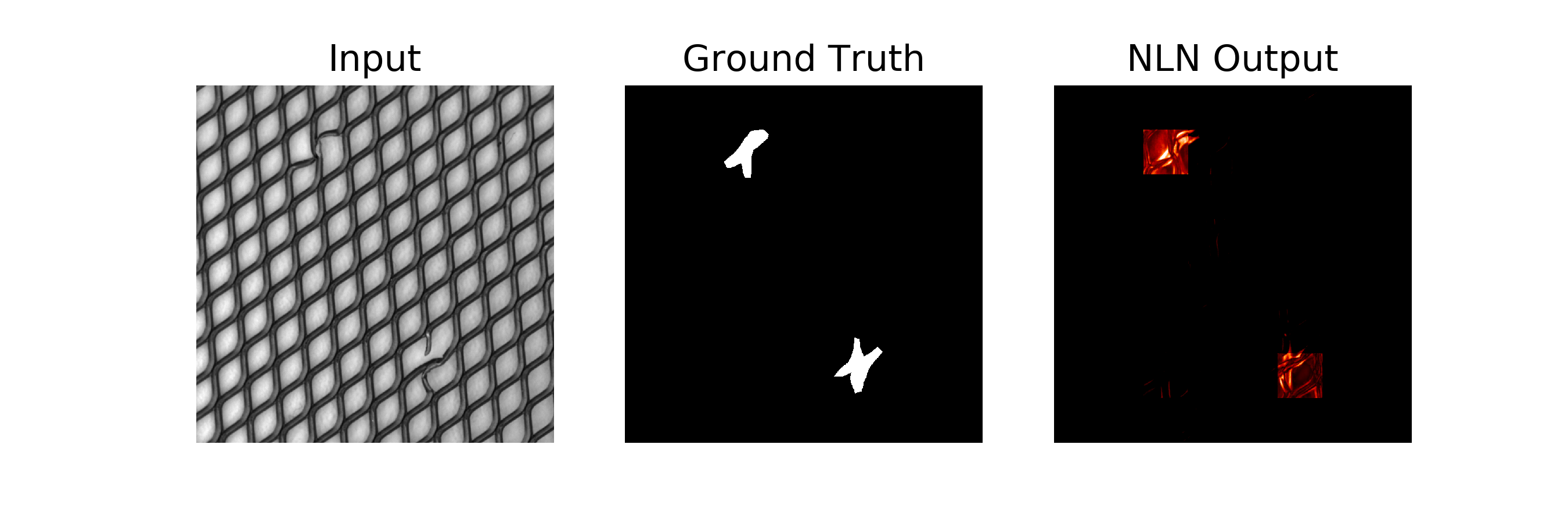}
                \caption{Grid}
                \label{fig:grid}
             \end{subfigure}

            \begin{subfigure}[b]{0.49\textwidth}
                \centering
                \includegraphics[width=\linewidth,trim={3cm 0 2cm 0},clip]{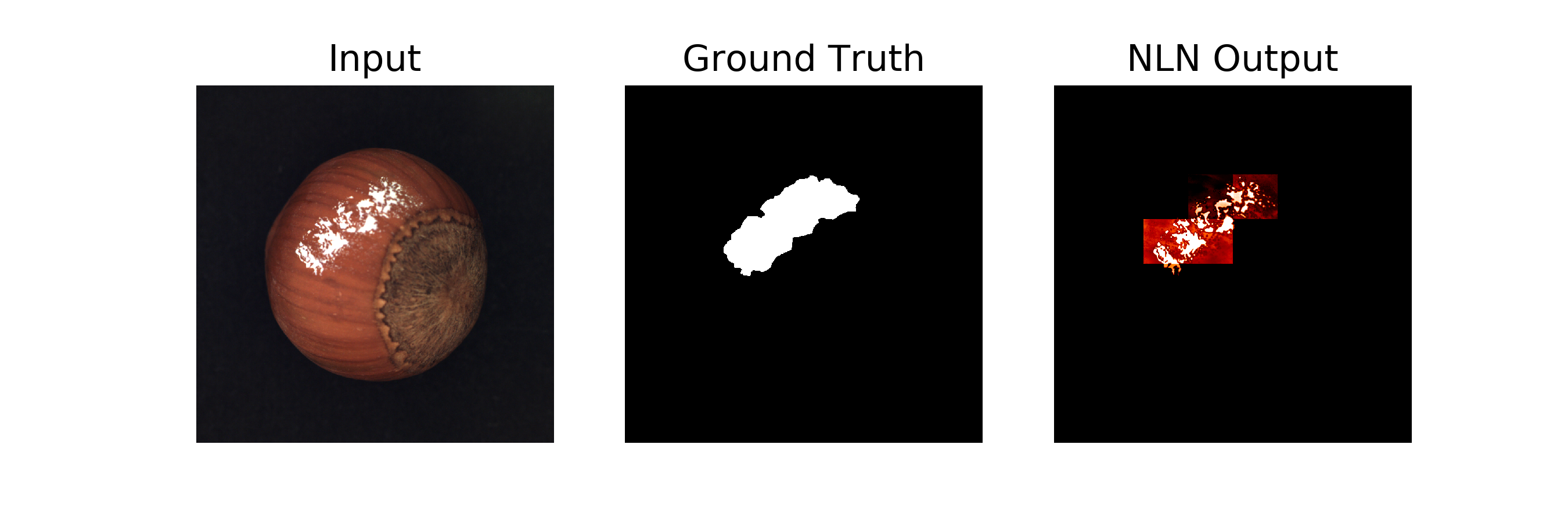}
                \caption{Hazelnut}
                \label{fig:hazelnut}
             \end{subfigure}
             \hfill
             \begin{subfigure}[b]{0.49\textwidth}
                 \centering
                 \includegraphics[width=\linewidth,trim={3cm 0 2cm 0},clip]{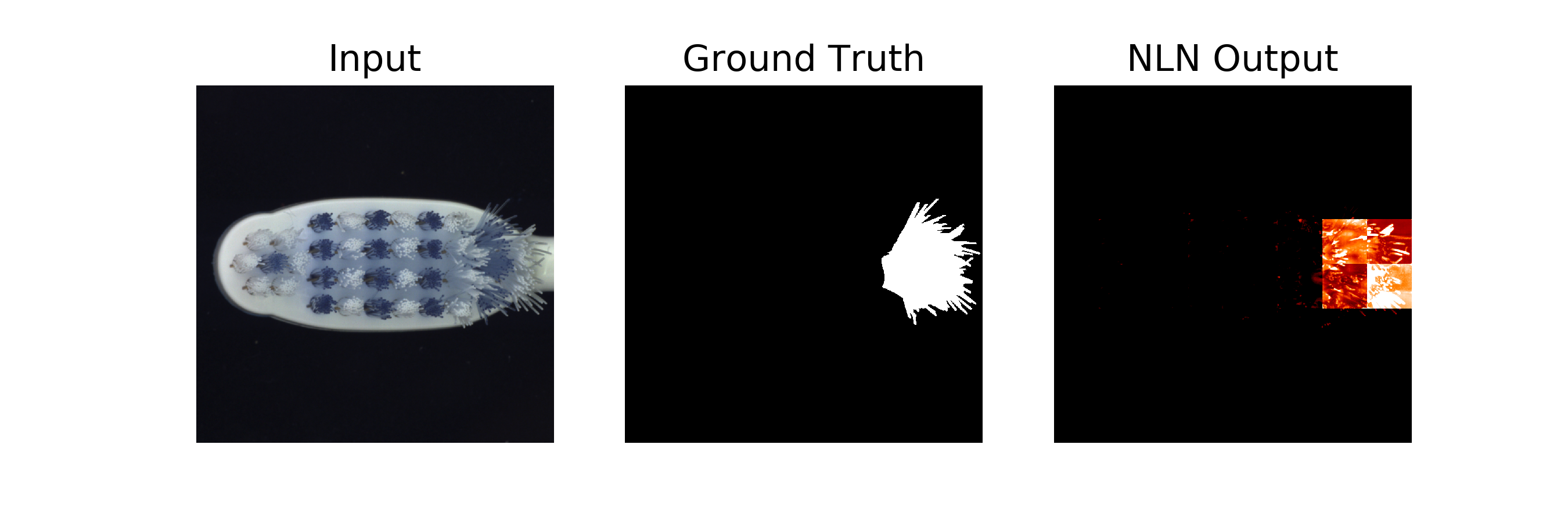}
                 \caption{Toothbrush}
                 \label{fig:toothbrush}
              \end{subfigure}
                 
            \caption{Pixel-level anomaly detection using NLN for four different MVTec-AD classes in the textures (top) and objects (bottom) categories.}
            \label{fig:segmentation}
        \end{figure*}
   
    \subsection{Evaluation methodology}
        To measure the performance of the NLN-enabled models, they are trained multiple times on a specific dataset, each time removing a different class or classes from the training set, thereby testing the novelty detection performance on every class present in a given dataset. We do this according to \cite{Burlina2019}, such that both the single-class or Single-Inlier-Multiple-Outlier (SIMO) and the multi-class or Multiple-Inliers-Single-Outlier (MISO) performance are evaluated. 
        
        We use the Area Under the Receiver Operating Characteristic (AUROC) score to evaluate and compare the performance of the NLN-algorithm. The AUROC metric measures the area under the ROC curve of true positive rates and false positive rates for different threshold values.  Furthermore, we evaluate the per-pixel detection performance of our NLN-enabled models using Intersection over Union (IoU) score. The IoU metric is a measure of the overlap between the predicted regions and their corresponding ground-truth.  
        
        We limit our evaluation to only autoencoders as we find comparison with methods that rely on SSL \cite{Yi2020,Chen2021,Sohn2020}, pretrained feature extractors \cite{Hendrycks2018, Tack2020,Sohn2020,Chen2021} or computationally expensive inference \cite{Yi2020} are not easily comparable on AUCROC alone across multiple datasets. It has been well documented that using pretrained feature extractors and SSL losses result in improved performance. However, they typically require orders of magnitude more parameters \cite{Bergmann2021}, and are not easily applicable across datasets or evaluation strategies. Furthermore, we regard the simplicity of AEs a crucial attribute. This is in contrast with the significant augmentation found in~\cite{Li2021} and the challenge of applying patch-dependant methods~\cite{Yi2020} to different datasets of varying resolutions and anomaly types. 
        
    \subsection{Datasets}
        We evaluate our work on four different datasets, namely MNIST \cite{LeCun2010}, CIFAR-10 \cite{Krizhevsky2009}, Fashion-MNIST \cite{Xiao2017} and MVTec-AD \cite{Bergmann2019a}. MNIST is a dataset consisting of $28\times28\times1$ handwritten digits between 0 and 9. The complexity of the dataset is low and therefore our method performs best on it. Similarly, Fashion-MNIST is composed of $28\times28\times1$ images of different types of articles of clothing. This dataset is used as an intermediary difficulty, between MNIST and CIFAR-10. CIFAR-10 is an object recognition dataset consisting of $32\times32\times3$ images of 10 different classes. It is the most challenging dataset for novelty detection as each of the semantic classes may appear at different scales, viewing angles and have changing backgrounds \cite{Ahmed2019}. The MVTec-AD dataset is an industrial anomaly detection dataset consisting of 15 different classes in 2   categories - objects and textures. The 10 object classes contain regularly positioned objects photographed in high resolution from the same viewing angle and the 5 texture classes contain repetitive patterns.  For training on the MVTec-AD dataset we follow the augmentation scheme proposed in \cite{Bergmann2019a}, where random rotations and crops are applied to the dataset that is broken into $128\times128$ patches. For more details about the dataset's composition and the augmentation performed see \cite{Bergmann2019a}.  
    
      \begin{table}[h!]
      \begin{minipage}[b]{\linewidth}
        \centering
         \resizebox{\linewidth}{!}{ \begin{tabular}{|c|c|c|c|c|c|}
                \hline
                &\textbf{AE} &\textbf{AAE} & \textbf{VAE} &\textbf{AE-res} & \textbf{AE-con} \\ 
                \hline
                \hline
                \textbf{MNIST}    & 9.80\%        & 17.65\%       & 11.65\%       & 10.13\%        & 14.18\% \\
                \textbf{CIFAR-10} & 6.92\%        & 7.41\%        & 1.29\%        & 7.30\%         & 6.68\%  \\
                \textbf{F-MNIST}   & 11.52\%      & 9.53\%        & 10.31\%       & 11.95\%        & 11.52\%  \\
                \hline
            \end{tabular}}
            \caption{Mean MISO AUROC percentage increase using NLN}
            \label{tab:MISO_performance_increase}
        \end{minipage}
        \newline 
         \begin{minipage}[b]{\linewidth}
            \centering
             \resizebox{\linewidth}{!}{    \begin{tabular}{|c|c|c|c|c|c|}
                \hline
                &\textbf{AE} &\textbf{AAE} & \textbf{VAE} &\textbf{AE-res} & \textbf{AE-con} \\ 
                \hline
                \hline
                \textbf{MNIST}     & 3.45\%   & 4.92\%       & 3.99\%       & 3.47\%        & 4.21\% \\
                \textbf{CIFAR-10}  & 7.65\%   & 8.81\%       & 6.36\%       & 6.89\%        & 8.02\%  \\
                \textbf{F-MNIST}   & 5.66\%   & 5.12\%       & 5.31\%       & 5.65\%        & 3.49\%  \\
                \hline
            \end{tabular}}
            \caption{Mean SIMO AUROC percentage increase using NLN }
            \label{tab:SIMO_performance_increase}
        \end{minipage}
        \end{table}

     \begin{table*}[h!]
        \centering
        \begin{minipage}[b]{0.42\linewidth}
        \resizebox{\linewidth}{!}{
        \begin{tabular}{|c|c|c|c|}
            \hline
             \textbf{Model} & \textbf{MNIST} & \textbf{CIFAR-10} & \textbf{F-MNIST} \\
             \hline
             \hline
              GANomaly~\cite{Akcay}          & 0.753 & 0.532 & 0.679\\
              Skip-GAN~\cite{Akcay2019}      & 0.492 & 0.629 & 0.515\\
              OC-GAN~\cite{Perera2019}       & 0.683 & 0.510 & 0.678\\
              VAE~\cite{An2015}              & 0.515 & 0.497 & 0.521 \\
              AnoGAN~\cite{Schlegl2017}      & 0.632 & 0.434 & 0.510 \\ 
              EGBAD~\cite{Zenati2018}        & 0.656 & 0.496 & 0.500 \\
              DKNN~\cite{Bergman2020}        & 0.791 & \textbf{0.714} & 0.746 \\
              \textbf{Ours}      & \textbf{0.921} &  0.560 & \textbf{0.763} \\
               \hline
        \end{tabular}}
        \caption{Mean MISO novelty detection AUROC, bold is best.}
        \label{tab:MISO}
        \end{minipage}
        ~~
        \begin{minipage}[b]{0.54\linewidth}
        \resizebox{\linewidth}{!}{
        \begin{tabular}{|c|c|c|c|c|}
            \hline
             \textbf{Model} & \textbf{MNIST} & \textbf{CIFAR-10} & \textbf{F-MNIST} & \textbf{MVTec-AD} \\
             \hline
             \hline
             GANomaly~\cite{Akcay}         & 0.965 & 0.695 & 0.906 & 0.762 \\
             OC-GAN~\cite{Perera2019}      & 0.975 & 0.657 & 0.924 & 0.756 \\
             AnoGAN~\cite{Schlegl2017}     & 0.912 & 0.618 & 0.817 & 0.600 \\ 
             LFD~\cite{Hong2020}           & 0.977 & - & 0.927 & 0.777\\
             CBiGAN~\cite{Carrara2021}     & - & - & - & 0.770 \\
             CAVGA-D$_u$~\cite{Venkataramanan2020}& \textbf{0.986} & 0.737 & 0.885 & - \\
             DKNN\footnotemark~\cite{Bergman2020}        & 0.917 & \textbf{0.890}  & 0.938 & 0.750 \\
             \textbf{Ours}      & 0.974  & 0.658 & \textbf{0.941} & \textbf{0.783}\\
             \hline
        \end{tabular}}
        \caption{Mean SIMO novelty detection AUROC, bold is best. }
        \label{tab:SIMO}
        \end{minipage}
        \end{table*}
        
     \subsection{Model and parameter selection}
        \label{subsec:params}
        In order to evaluate our work across a number of different datasets we adapt our models accordingly. We adopt autoencoding the architecture specified in \cite{Bergmann2019a} for the evaluation of the NLN algorithm on the MVTec-AD dataset. For MNIST, CIFAR-10 and F-MNIST we modify a LeNet \cite{Lecun1998a} based autoencoding architecture. The encoder consists of 3 convolutional layers and the decoder has 3 transposed-convolutional layers. A base number of filters of 32 is used for the AE and is increased or decreased on each subsequent layer by a factor of 2.  We use \textit{ReLU} activations for all models and they are trained for 50 epochs using ADAM \cite{Kingma2015} with a learning rate of $1\times10^{-4}$. The image-based discriminators $d_{\vb{x}}$ use the same architecture as the encoder, except the final layer, which is a dense layer with a \textit{sigmoid} activation. The latent discriminator for the AAE consists of 3 dense layers with \textit{Leaky ReLU} activations and a dropout rate of $0.3$. The base layer size is 64 and is increased by a factor of 2 for each subsequent layer. Furthermore, we treat the maximum number of neighbours, $K$, the latent dimensionality, $L$, and the NLN contribution, $\alpha$, as hyper-parameters of our algorithm.

    \subsection{Results}
        We evaluate the performance increase of the NLN algorithm for a variety of autoencoding models across a number of different datasets in both the MISO-context in Table~\ref{tab:MISO_performance_increase} and SIMO-context in Table~\ref{tab:MISO_performance_increase}. Here the best performing reconstruction error-based AUROC is compared with the best performing NLN-enabled model for each architecture. The NLN-based AEs achieve a performance increase between 17\% and 1\% across the three MISO-datasets and 8\% and 3\% for the SIMO-case. We suspect the low performance gains in the SIMO-case of the NLN-enabled AEs are due there being fewer latent neighbours to select from, thereby reducing performance.   
        
        In Table~\ref{tab:MISO} we present the MISO-based class-averaged AUROC comparison of autoencoding models. For MNIST, the optimal configuration is a feature consistent AE with $K=2$, $L=32$ and $\alpha=1.0$, for CIFAR-10 we use the discriminative AE when $K=1$, $L=32$ and $\alpha=0.5$. Finally for F-MNIST, we use a discriminative AE when $K=1$, $LD=64$ and $\alpha=0.9$. Here we see that the NLN-algorithm gives significant performance increases for MNIST and F-MNIST, even above the pretrained ResNet-50 proposed by \cite{Bergman2020}. Furthermore, we see that OCGAN~\cite{Perera2019} is not performant in a MISO context, this indicates that our NLN algorithm may offer a more robust solution to the generalisability problem in AEs. We show that AEs do not perform particularly well on CIFAR-10. This is expected, as images from the same class in contain substantially different pixel-level information. For example the \textit{aeroplane} class contains images of both the cockpit of a grounded Boeing 747 as well a fighter-jet photographed from the side-view in mid-flight. In effect, the MSE between non-novel images in the same class, can be greater than novel images thereby reducing the efficacy of MSE based novelty detectors on CIFAR-10. 
        
        We present the class-averaged AUROC scores for the SIMO-based evaluation in Table~\ref{tab:SIMO}. Here the optimal method for MNIST is a discriminative AE, with $LD=128$, $K=3$ and $\alpha=1.0$ and for CIFAR-10 we find the optimal method to be a vanilla AE with $LD=256$, $K=1$ and $\alpha=0.75$. Furthermore, we find the best performing method on F-MNIST to be a VAE with $LD=32$, $K=3$ and $\alpha=0.9$. For the MVTec-AD dataset we use a discriminative AE with $LD=128$, $K=1$ and $\alpha=0.8$. It is clear that the attention guided VAE (CAVAGA)~\cite{Venkataramanan2020} method performs best on MNIST whereas DKNN~\cite{Bergman2020} on CIFAR-10. However, it is evident that the NLN-enabled autoencoding models offer increased performance over existing autoencoding and ResNet-based architectures for both the F-MNIST and MVTec-AD datasets in the SIMO context. 
        
\begin{figure}[h!]
            \centering
            \includegraphics[width=0.975\linewidth]{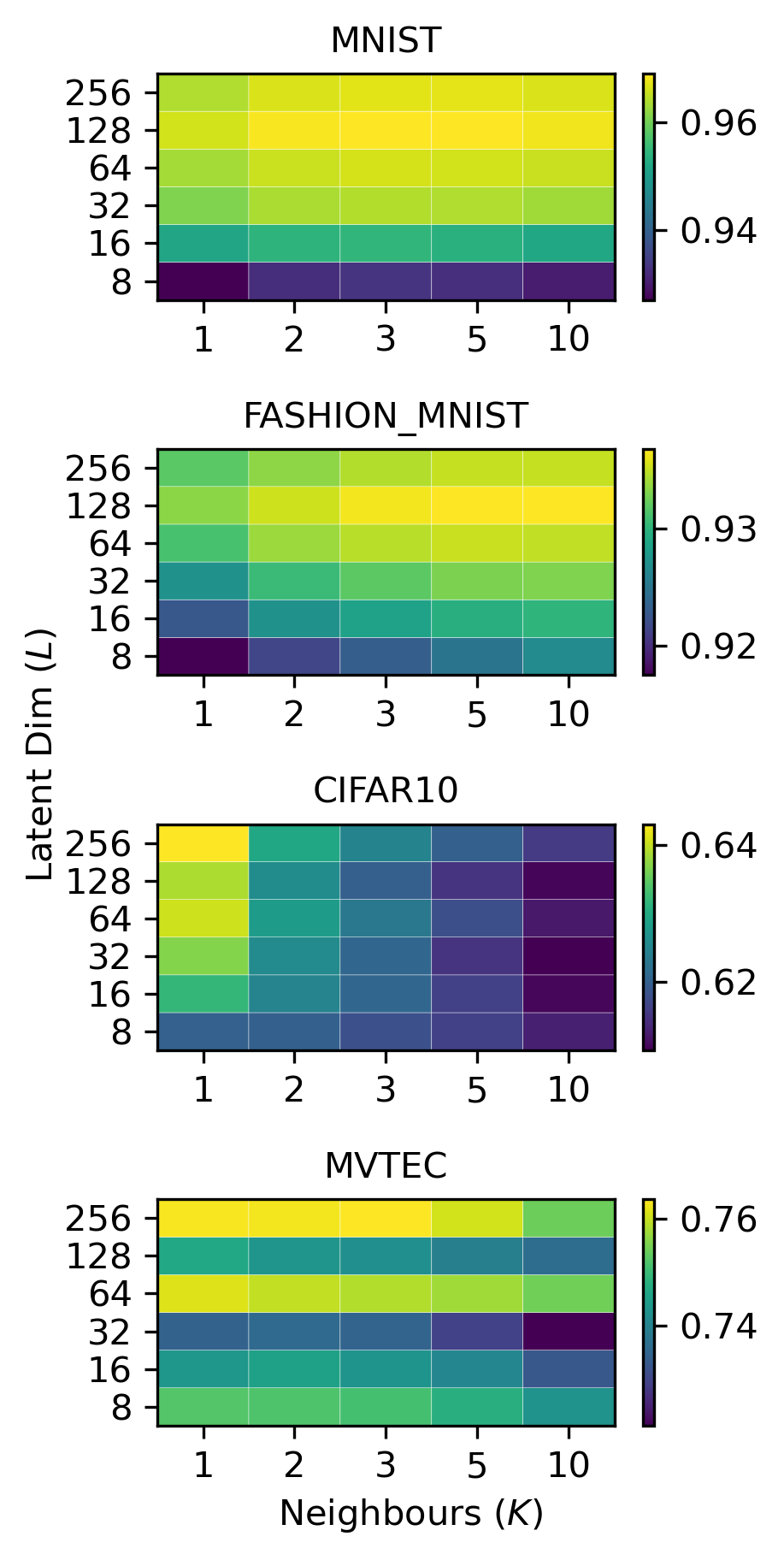}
            \caption{Vanilla Autoencoder AUROC sensitivity to number of neighbours and latent dimensions in SIMO-context for $\alpha=0.8$}
            \label{fig:sensitivity}
        \end{figure}

        In Figure~\ref{fig:sensitivity} we show the effect of varying $L$ and $K$ on AUROC scores for vanilla AE in the SIMO context when $\alpha=0.8$. For F-MNIST and MNIST a maximum AUROC score is found for $L=128$ and $K>3$, whereas for CIFAR-10 the optimal is found when $L=256$ and $K=1$. Finally it is shown that the vanilla AE offers best image-based AUROC performance when $L=256$ and $K=3$.

        \footnotetext{We use the authors implementation for all datasets other than MVTec-AD, here we use our own Tensorflow-based implementation}

        \begin{table*}[h!]
        \centering
%        \begin{minipage}[b]{0.49\linewidth}
%        \resizebox{\linewidth}{!}{
        \begin{tabular}{|l|l|c|c|c|c|c|}
            \hline
%            \rowcolor{lightgray} 
             &\textbf{Class} & \textbf{AE-L2 \cite{Bergmann2019a}} & \textbf{AE-SSIM~\cite{Bergmann2019a}}  & \textbf{SMAI L2 \cite{Li2020} } &\textbf{VE-VAE\cite{Liu2020a}} &\textbf{Ours} \\
            \hline
            \hline
            %\multirow{5}{*}{Multirow}
             \parbox[t]{2mm}{\multirow{5}{*}{\rotatebox[origin=c]{90}{Textures}}}
              
              &  carpet &   0.59    &  0.87 & \textbf{0.88}  &   0.78      & 0.82      \\ %\hline
              &grid     &   0.90    &  0.94 &  \textbf{0.97} &   0.73      & 0.86      \\ %\hline
              &leather  &   0.75    &  0.78 &  0.86   &  \textbf{ 0.95}    & 0.85     \\ %\hline
              &tile    &   0.51    &  0.59 &  0.62   &   \textbf{0.80}     & 0.51     \\ %\hline
              &wood     &   0.73    &  0.73 & \textbf{0.80}   &   0.77     & 0.72      \\ \hline
              &mean     &   0.70    &  0.78 &  \textbf{0.83}   &   0.81    & 0.75     \\ %\hline
              \hline
              \hline
             \parbox[t]{2mm}{\multirow{10}{*}{\rotatebox[origin=c]{90}{Objects}}}
              &bottle    &   0.86    &  0.93    &  0.86   &  0.87    & \textbf{0.95}   \\ %\hline
              &cable     &   0.86    &  0.82    &  \textbf{0.92}   &  0.90    & 0.90   \\ %\hline
              &capsule   &   0.88    &  \textbf{0.94}    &  0.93   &  0.74    & \textbf{0.94}   \\ %\hline
              &hazelnut  &   0.95    &  0.97    &  0.97   &  \textbf{0.98}    & \textbf{0.98}   \\ %\hline
              &metal nut &   0.86    &  0.89    &  0.92   &  \textbf{0.94}    & 0.88   \\ %\hline
              &pill      &   0.85    &  0.91    &  \textbf{0.92}   &  0.83    & \textbf{0.92}   \\ %\hline
              &screw     &   0.96    &  0.96    &  0.96   &  \textbf{0.97}    & \textbf{0.97}   \\ %\hline
              &toothbrush&   0.93    &  0.92    &  0.96   &  0.94    & \textbf{0.97}    \\ %\hline
              &transistor&   0.86    &  0.90    &  0.85   &  \textbf{0.93}    & 0.85   \\ %\hline
              &zipper    &   0.77    &  0.88    &  0.9    &  0.78    & \textbf{0.96}   \\ \hline
              &mean      &   0.88    &  0.91    &  0.92   &  0.89    & \textbf{0.93}  \\ %\hline
              \hline
        \end{tabular}%}
        \caption{Pixel-based novelty detection (Segmentation) AUROC score for autoencoding models, where bold is best.} 
        \label{tab:MVTEC_seg}
        \end{table*}
        
    We evaluate the pixel-level anomaly detection performance in Table~\ref{tab:MVTEC_seg}, and illustrate the model outputs in Figure~\ref{fig:segmentation} of both texture and object classes. In all cases we use a vanilla AE with $K=1$, $L=128$ and $\alpha=0.6$. It is clear that the NLN-enabled AE demonstrates performance increases in the object classes of MVTec-AD. However, this is not the case for the texture classes. We suspect that this is due to our NLN-enabled AE not being able to distinguish between different texture-patches. This behaviour is similarly demonstrated in \cite{Bergmann2019a}, and we believe that this is an inherent weakness of standard autoencoding architectures. 
        
     \begin{figure*}[h!]
        \centering
        \includegraphics[width=\linewidth]{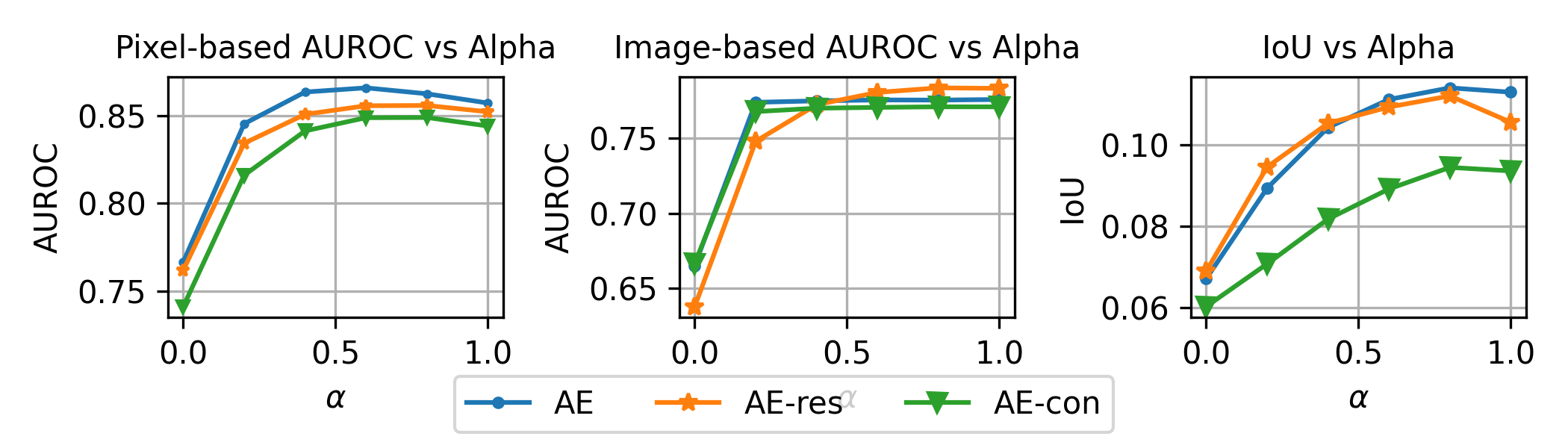}
        \caption{AUROC and IoU sensitivity to varying $\alpha$ of the NLN-enabled autoencoding models applied to the MVTec-AD dataset.}
        \label{fig:alpha_sensitivity}
    \end{figure*}

    In Figure~\ref{fig:alpha_sensitivity} we illustrate the effect on varying alpha for the NLN-enabled autoencoding models used for the MVTec-AD dataset. Here it is demonstrated, that the NLN-based model obtain optimal AUROC segmentation-performance when $0.25<\alpha<0.8$, whereas to optimal AUROC detection-performance occurs when $\alpha>0.6$. Finally we illustrate that the optimal IoU value is obtained at $\alpha=0.8$, thus demonstrating the benefit of including the reconstructions of nearest-neighbours in the calculation of the anomaly score.    
     
    \subsection{Time and memory efficiency}
    \label{sec:complexity}
    The NLN-algorithm requires a forward pass through an encoder, a KNN search of the latent-space generated by the training samples, and a forward pass of a given point's nearest neighbours through a decoder. We evaluate the models on a Nvidia T4, where a forward pass of a single image from the MVTec-AD dataset takes 7.41~ms for the encoder and 9.63~ms for the decoder. In comparison, a ResNet50 used in \cite{Bergman2020,Bergmann2020,Perera2019a} requires 43.3ms for a forward pass of a single image. This means that our method is between 1.3$\times$ and 2.5$\times$ more efficient for a forward pass, depending on the architecture used.   

    For the KNN search we use a k-d tree implementation of the KNN search, which has a inference time complexity of $\mathcal{O}(K L \log{N})$. Where $K$ is the number of neighbours, $L$ is the latent dimensionality and $N$ is the number of points in the training set. In the case of the NLN-enabled models presented in this work, we find a latent dimensionality of 128 sufficient, whereas the ResNet50 in \cite{Bergman2020} uses 2048 dimensional latent space. This means that our work offers a $16\times$ reduction in KNN search inference time incomparison with \cite{Bergman2020}.  
    
    Finally, our method has comparable storage requirements as other AE based models \cite{Bergmann2021} in terms of number of trainable parameters. For comparison, the AE-con model used for MVTec-AD has 1.79 million parameters, whereas the ResNet-50 from \cite{Bergman2020} has 25.58 million parameters. The only storage-based overhead of the NLN-algorithm is the requirement of amortising the embeddings of the training set as suggested in \cite{Bergman2020}. In the case of the bottle-class of the MVTec-AD dataset, there is an additional storage requirement of 6.85~MB\footnote{$209\text{~images} \times 16 \times 4\text{~augmented patches} \times 128 \text{~latent dimensions}  \times 32 \text{~bits}=6.85\text{MB}$ of additional memory}
    
    \section{Ablation study}
The AUROC performance of the NLN-algorithm is demonstrated in Table~\ref{tab:loss_ablation} when the loss function varied. The term in the first column, $\mathcal{L}_{\text{recon}}$, represents the standard reconstruction error given by Equation~\ref{eq:orig} and $\mathcal{L}_{\text{NLN}}$ shows the NLN-based reconstruction loss given in the first half of Equation~\ref{eq:nln}. $\mathcal{L}_{\text{con}}$ represents the feature consistent adaption given by the first half of Equation~\ref{eq:nln_con} and  $\mathcal{L}_{\text{total}}$ is equivalent to the score obtained from Equation~\ref{eq:nln_con}. It can be seen that through the utilisation of all terms in NLN-loss formulation we obtain optimal performance. 

 \begin{table}[h!]
            \centering
            \begin{tabular}{|c|c|c|c|c|c|}
                \hline
                 \textbf{Dataset} & \textbf{$\mathcal{L}_{\text{recon}}$} & \textbf{$\mathcal{L}_{\text{NLN}}$} & \textbf{$\mathcal{L}_{\text{con}}$} & \textbf{$\mathcal{L}_{\text{total}}$} \\
                 \hline
                 \hline
                  MNIST            & 0.778 & 0.822 & 0.913 & 0.921 \\
                  FMNIST           & 0.669 & 0.702 & 0.719 & 0.738 \\
                  CIFAR-10         & 0.511 & 0.513 & 0.551 & 0.553  \\
                 \hline
            \end{tabular}
            \caption{MISO AUROC performance of AE-con for different losses terms when $K=2$, $L=32$ and $\alpha=0.9$ }
            \label{tab:loss_ablation}
\end{table}

     \section{Discussion and conclusions}
 \label{sec:conclusion}
 Autoencoders learn to generalise to unseen classes which is a problem when they are used for novelty detection. In this work, we demonstrate that when the reconstructions of a model's nearest-latent-neighbours are harnessed we can more effectively and efficiently mitigate this problem in comparison with the state-of-the-art. This is achieved through a fairly simple algorithm that is agnostic to both the AE's architecture and its error method. We experimentally prove that the addition of the NLN algorithm consistently yields performance increases for various autoencoding architectures and various datasets and is competitive with the state-of-the-art autoencoding models. This is achieved without complex augmentation, using pretrained networks or computationally expensive inference. We note that the complexity of CIFAR-10 and the texture classes of MVTec-AD result in modest performance, but we expect this can be solved using more robust error functions or using SSL to obtain even better latent representations.  
    \section*{Acknowledgements}
This work is part of the \textit{Perspectief} research programme Efficient Deep Learning (\href{https://efficientdeeplearning.nl/}{EDL}), which is financed by the Dutch Research Council (NWO) domain Applied and Engineering Sciences (TTW).
    \bibliography{references.bib}
\end{document}